%% file: main.tex
\documentclass[letterpaper, 10 pt, conference]{ieeeconf}
\IEEEoverridecommandlockouts
\usepackage{graphicx}
\usepackage{caption}
\usepackage{subcaption}
\usepackage{multirow}
\usepackage{tabularx}
\usepackage{caption}
\usepackage{amsmath,amssymb}
\usepackage[dvipsnames,table,xcdraw, HTML]{xcolor}
\usepackage{hyperref}
\hypersetup{
    colorlinks=true,
    linkcolor=orange,
    filecolor=magenta,      
    urlcolor=orange,
    citecolor=orange,
}
\usepackage[capitalise, nameinlink]{cleveref}
\makeatletter
\let\NAT@parse\undefined
\makeatother
\usepackage[numbers,sort,compress]{natbib}
\usepackage{color, colortbl, soul}
\usepackage{multicol}
\usepackage{listings} 
\usepackage{tcolorbox}
\let\labelindent\relax
\usepackage{enumitem}
\usepackage{booktabs}
\usepackage{siunitx}
\usepackage{amsfonts} 

\usepackage{algorithm}
\usepackage{algpseudocode}
\usepackage[ruled,vlined,linesnumbered,algo2e]{algorithm2e}

\DeclareMathOperator*{\argmin}{arg\,min}

\newcommand{\rhyme}[0]{\texttt{RHyME} }
\usepackage{xcolor}
\usepackage{soul}  
\usepackage{color, soul}
\usepackage[customcolors]{hf-tikz}
\usepackage{dsfont} 
\usepackage{colortbl}
\usepackage{multirow}
\usepackage{transparent}

\definecolor{lightgreen}{rgb}{0.8,1,0.8}

\definecolor{perfblue}{RGB}{64, 114, 175}
\newcommand{\algcommentlight}[1]{\textcolor{perfblue}{\transparent{0.8}\small{\texttt{\textbf{//\hspace{2pt}#1}}}}} 

\DeclareSymbolFont{rsfs}{U}{rsfs}{m}{n}
\DeclareSymbolFontAlphabet{\mathscrsfs}{rsfs}
\DeclareMathAlphabet{\mathcal}{OMS}{cmsy}{m}{n}

\title{\LARGE \bf One-Shot Imitation under Mismatched Execution }

\author{Kushal Kedia$^{*,1}$, Prithwish Dan$^{*,1}$, Angela Chao$^{1}$, Maximus A. Pace$^{1}$, Sanjiban Choudhury$^{1}$
\thanks{$^{*}$Equal Contribution}
\thanks{$^{1}$Department of Computer Science, Cornell University}
}%

\begin{document}


\twocolumn[{%
\renewcommand\twocolumn[1][]{#1}%
\maketitle
\begin{center}
\captionsetup{type=figure}
\vspace{-0.5cm}
\includegraphics[width=1.0\textwidth]{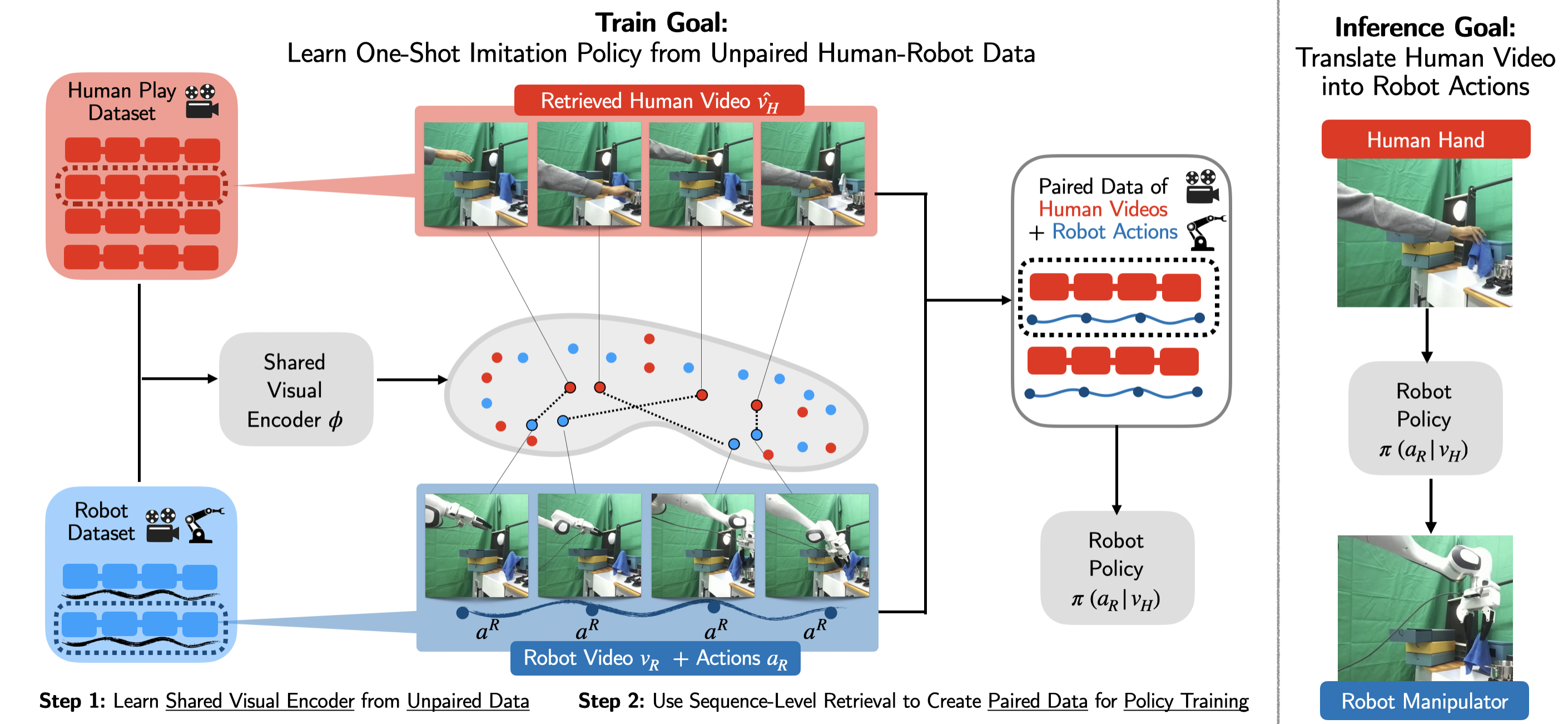}
 \captionof{figure}{\textbf{Overview of RHyME.} We introduce \texttt{RHyME}, a hierarchical framework that trains a robot policy to mimic a long-horizon video from a demonstrator that exhibits mismatched task execution. \textbf{Train Goal (Left):} Given unpaired human-robot datasets, \texttt{RHyME} ``imagines" employs sequence-level similarity functions to create a paired dataset for training one-shot imitation robot policies. \textbf{Inference Goal (Right):} Our robot policy translates a human video into robot actions to perform the specified long-horizon task. }
    \label{fig:introduction}
\end{center}
}]
\renewcommand{\thefootnote}{}
\footnote{\textsuperscript{1}Cornell University \textsuperscript{*}Equal Contribution}

\thispagestyle{empty}
\pagestyle{empty}

\begin{abstract}
Human demonstrations as prompts are a powerful way to program robots to do long-horizon manipulation tasks.
However, translating these demonstrations into robot-executable actions presents significant challenges due to execution mismatches in movement styles and physical capabilities. Existing methods for human-robot translation either depend on paired data, which is infeasible to scale, or rely heavily on frame-level visual similarities that often break down in practice. To address these challenges, we propose \texttt{RHyME}, a novel framework that automatically pairs human and robot trajectories using sequence-level optimal transport cost functions. Given long-horizon robot demonstrations, \texttt{RHyME} synthesizes semantically equivalent human videos by retrieving and composing short-horizon human clips. This approach facilitates effective policy training without the need for paired data. \texttt{RHyME} successfully imitates a range of cross-embodiment demonstrators, both in simulation and with a real human hand, achieving over 50\% increase in task success compared to previous methods. We release our code and datasets at \href{https://portal-cornell.github.io/rhyme/}{this website}.
\end{abstract}

\input{paper/introduction}
\input{paper/related_work}

\input{paper/problem}
\input{paper/approach}
\input{paper/experiments}
\input{paper/discussion}
\section{Acknowledgements}
Sanjiban Choudhury is supported in part by Google Faculty Research Award, OpenAI SuperAlignment Grant, ONR Young Investigator Award, NSF RI \#2312956, and NSF FRR \#2327973.

\bibliographystyle{IEEEtran}
\bibliography{IEEEabrv,refs}

\end{document}

%% file: paper/introduction.tex
\section{Introduction}

Human demonstrations offer an effective approach for programming robots to execute long-horizon manipulation tasks~\cite{Zakka2021XIRLCI, kumar2022inverse, bahl2022human, xu2023xskill}. Unlike language instructions, demonstrations are grounded in the task environment, providing rich cues for what steps to follow, which objects to interact with, and how to interact with them~\cite{bahl2023affordances, wang2023mimicplay}.

We view this as a translation problem where a human video must be translated into a series of robot actions~\cite{Jang2022BCZZT, Jain2024Vid2RobotEV, AVID, ChaneSane2023LearningVP}. However, training such policies typically requires paired human-robot demonstrations, which is impractical to collect at scale for long-horizon tasks. Although large-scale human videos (e.g., YouTube) and robot datasets exist~\cite{khazatsky2024droid, padalkar2023open}, they are unpaired, making them unsuitable for directly learning this translation.

Prior works leverage unpaired human and robot demonstrations to learn visual representations that map both human and robot images into a shared embedding space~\cite{xu2023xskill, wang2023mimicplay,Nair2022R3MAU, ma2022vip, xiao2022masked}. A policy is then trained to generate actions conditioned on robot video embeddings, and directly transferred at test time to work with embeddings from a human prompt video. However, a key assumption these works rely on is that the human and robot perform tasks with \emph{matched execution}, i.e., the human executes tasks in a visually similar way to that of the robot (e.g. slowly moving one arm with a simple grasp). 
In reality, humans often act more swiftly, use both hands for manipulation, or even execute multiple tasks simultaneously, creating a mismatch in execution styles. This mismatch leads to misalignment between the human and robot embeddings, hindering direct policy transfer.

We tackle this problem of imitation under \emph{mismatched execution}. \textbf{\emph{Our key insight is that while a human and robot may perform the same task in visually and physically different ways, we can establish a high-level equivalence by reasoning over the entire sequence of image embeddings they generate.}} We show that while individual image embeddings may appear different between human and robot, we can construct sequence-level similarity functions where the two are closer. Notably, we can do this \textbf{\emph{without fine-tuning representations}} on paired data.

We propose \texttt{RHyME} (\texttt{R}etrieval for \texttt{Hy}brid Imitation under \texttt{M}ismatched \texttt{E}xecution), a framework which trains a robot policy to follow a long-horizon demonstration from a mismatched expert without access to paired human-robot videos (Fig.~\ref{fig:introduction}). First, \texttt{RHyME} defines a sequence-level similarity metric between human and robot embeddings, using optimal transport to measure alignment. Given a robot trajectory and a database of human play data, \texttt{RHyME} \textit{imagines} a long-horizon video by retrieving and composing short-horizon snippets of human demonstrations similar to the robot video. This retrieval process is guided by an optimal transport similarity metric between human and robot sequences. The framework trains a policy using a hybrid approach, incorporating both real robot demonstrations and imagined human sequences.  Our contributions are:
\input{paper/figures/intro_fig}

\begin{enumerate}[leftmargin=*]
\itemsep0em 
\item We introduce \texttt{RHyME}, a retrieval-based algorithm to create paired datasets of human and robot trajectories, enabling one-shot imitation under mismatched execution.
    \item We propose three novel datasets for cross-embodiment learning in the Franka Kitchen simulator with progressively increasing execution mismatch relative to the robot.
    \item In both simulation and the real world, we demonstrate that \texttt{RHyME} outperforms baselines, increase task completion rates $2$x when imitating unseen human videos. Additionally, we show that sequence-level similarity functions significantly improve retrieval performance compared to frame-level similarity functions.
\end{enumerate}

%% file: paper/figures/intro_fig.tex

%% file: paper/related_work.tex
\vspace{-1mm}
\section{Related Work}

Human demonstrations have been utilized to guide robot manipulation policies in several different ways. We place our work relative to each cluster of related research.

\textbf{Tracking Reference Motion.} The imitation challenge is reduced to motion tracking when the robot receives the demonstrator's motion as input. Robots with human-like joint configurations can directly mimic human trajectories, a method applied to humanoid robots with similar morphologies~\cite{Pollard2002AdaptingHM,Nakaoka2003GeneratingWB,Kim2009StableWM,Suleiman2008OnHM,Koenemann2014RealtimeIO,Peng2018SFVRL} and robotic hands~\cite{Handa2019DexPilotVT,GarciaHernando2020PhysicsBasedDM,Qin2021DexMVIL,mandikal2021dexvip,Ye2022LearningCG,Shaw2022VideoDexLD,telekinesis,arunachalam2022dime}. When the execution capabilities are mismatched, either human models are simplified~\cite{Pollard2002AdaptingHM,Kim2009StableWM,Koenemann2014RealtimeIO} or robot trajectories~\cite{Nakaoka2003GeneratingWB,Suleiman2008OnHM,kuang2024ram,bharadhwaj2024track2act} are optimized to approximately match the reference trajectory.~\cite{Peng2018SFVRL} shows a method to extract human poses from videos and use that as a reference for training reinforcement learning agents. Similarly, human demonstrations have been used to guide reinforcement learning for robotic hands \cite{Handa2019DexPilotVT, GarciaHernando2020PhysicsBasedDM,Qin2021DexMVIL,mandikal2021dexvip}. More recently, large-scale video datasets of humans on the internet have been used to extract hand positions and adapted for robot manipulation~\cite{telekinesis,Shaw2022VideoDexLD, arunachalam2022dime, Ye2022LearningCG, zhu2024vision}, or rely on other common abstractions such as optical flow as a common trajectory representation across embodiments \cite{xu2024flow}, still requiring execution to be matched. Distinct from these works, we focus on learning robot manipulation tasks directly from human RGB videos without explicit motion input.

\textbf{Learning Reward Functions from Demonstrator Videos. } These methods tackle the problem of matching the demonstrator behaviors when their motion cannot be simply mimicked. Still, their videos contain useful task information and can be used to learn reward functions for reinforcement learning on the robot. A general method in this line of research~\cite{Nair2017CombiningSL, Sieb2019GraphStructuredVI, schmeckpeper2020learning, kumar2022inverse} is to extract reward functions that encourage the robot to manipulate objects in the same way as the video. For example,~\cite{Nair2017CombiningSL} self-supervises the robot by making it learn to match the demonstrations of a demonstrator perturbing a rope. GraphIRL~\cite{kumar2022inverse} extracts sequences of object pose movement from the demonstrator and enforces temporal cyclic consistency with the robot's execution. Other approaches frame the problem as task matching~\cite{Shao2020Concept2RobotLM, Chen2021LearningGR}, where the robot is rewarded when it is deemed to perform the same skill as the demonstrator. While these papers tackle the problem of mismatched execution directly, they require reinforcement learning to train robot policies, which is challenging for complex tasks.

\textbf{Learning Aligned Human-Robot Representations.} Another strategy for addressing this challenge is to train representations for both the robot and the human that are indistinguishable when performing the same task. This approach often frames the task as a video translation challenge, where the demonstrator's video is converted into a robot's perspective to simplify task mimicry \cite{ChaneSane2023LearningVP, AVID, Xiong2021LearningBW}. Additionally, methods like WHIRL \cite{bahl2022human} align videos by effectively masking out both the robot and the human, creating a neutral visual field. Embeddings can be aligned using datasets that either directly pair human and robot actions or utilize human preference datasets to rank image frames, as shown in X-IRL \cite{Zakka2021XIRLCI} and RAPL \cite{Tian2023WhatMT}. Unlike these methods, our approach does not depend on labeled human-robot correspondences.

\textbf{One-shot Visual Imitation from Demonstration Videos.}
We tackle this problem setting in our work where the robot imitates actions from human demonstration videos in a one-shot setting, i.e., the robot uses a prompt video as a guide, aiming to replicate the demonstrated actions after viewing them once~\cite{Jang2022BCZZT, Jain2024Vid2RobotEV,xu2023xskill, wang2023mimicplay, Duan2017OneShotIL, Dasari2020TransformersFO, Mandi2021TowardsMG, zhu2024vision}. If a paired dataset of human and robot videos executing the same task exists, the robot can learn to translate a prompt video into actions directly~\cite{Jang2022BCZZT, Jain2024Vid2RobotEV}. The closest to our work is the setting without paired data of human and robot skills. However, these works~\cite{wang2023mimicplay, xu2023xskill} train policies conditioning on robot videos and rely on zero-shot transfer to a prompt demonstration at test time. For example, XSkill~\cite{xu2023xskill} uses a self-supervised clustering algorithm based on visual similarity to align representations of human and robot videos. However, such an approach can falter when there are significant mismatches in execution. We address this issue by posing the visual imitation problem as a train-time retrieval problem. During training, we match robot videos to the closest human snippets from an unpaired play dataset to imagine synthetic demonstration videos. Training robot policies conditioned on these synthetic videos enable the robot to translate demonstration videos into robot actions.

%% file: paper/problem.tex
\section{Problem Formulation}
\textbf{Inference Time: Translate Human Demonstration Video to Robot Actions}. 
The robot's goal is to replicate a series of tasks demonstrated in a video using a policy $\pi(a_R | s_R, \mathbf{v_H})$ that translates the video into robot actions $a_R$ at state $s_R$. The human demonstration video is a sequence of images $\mathbf{v_H} = \{v_H^0, v_H^1, \dots, v_H^T\}$, where $T$ is video length.

\textbf{Train Time: Learning from Unpaired Human and Robot Data.} 
While training a policy with paired human-robot data is feasible, collecting such data at scale is impractical. Instead, we frame the problem as learning aligned embeddings from unpaired data, enabling the transfer of policies trained on robot embeddings to human embeddings.

We assume access to two datasets \textemdash a \textit{robot dataset} ($D_{\rm robot}$) of long-horizon manipulation tasks and a \textit{play dataset} ($D_{\rm play}$) of short-horizon human video clips showing interactions with objects and the environment.
The robot dataset, $D_{\rm robot} = \{(\mathbf{\xi_R}, \mathbf{v_R})\}$, comprises pairs of state-action trajectories and robot videos. Each robot trajectory, $\mathbf{\xi_R} = \{(s_R^0, a_R^0), (s_R^1, a_R^1), \dots, (s_R^T, a_R^T)\}$, represents the sequence of robot states and actions throughout an episode. Correspondingly, the video $\mathbf{v_R} = \{v_R^0, v_R^1, \dots, v_R^T\}$ is a sequence of images of the robot executing the task. 
The play dataset, $D_{\rm play} = \{\mathbf{v_H}\}$, consists of human videos that do not have direct correspondences with the robot dataset.
At test time, the demonstrator's video contains a set of tasks whose composition is unseen by the robot during training.
However, consistent with prior work~\cite{xu2023xskill}, we assume that the constituent tasks are individually covered both in $D_{\rm play}$ and $D_{\rm robot}$.

Our goal is to train two modules: a vision encoder and a robot policy. The vision encoder maps both human and robot videos into a shared embedding space to enable translation. We employ a video encoder $\phi(\mathbf{v})$ to extract a sequence of embeddings $\mathbf{z} = \{z_0, z_1, \dots, z_T\}$ for each frame from all videos\footnote{We encode a 1-timestep sliding window of 8 neighboring images to generate each image embedding.}. Then, given a human demonstration video  $\mathbf{v_H}$, we generate a sequence of latent embeddings $\mathbf{z_H}$. We aim to train a policy that conditions on the sequence of embeddings to predict robot actions $\pi(a_R | s_R, \mathbf{z_H})$ without access to paired human and robot data. We discuss how to train both the encoder and the policy in Section~\ref{sec:approach}.

%% file: paper/approach.tex
\vspace{-2mm}
\section{Approach}
\label{sec:approach}
\vspace{-1mm}
We present \texttt{RHyME}, a one-shot imitation learning algorithm that translates human videos into robot actions, without paired data. Before policy training, we first train a video encoder using a dataset of unpaired human and robot videos (Section ~\ref{sec:representation}). Then, this trained video-encoder is frozen and utilized for retrieval during policy training (Section~\ref{sec:opt}). At train time, given just a robot trajectory, \rhyme imagines a corresponding demonstration by retrieving and composing short-horizon human snippets. It then trains a policy to predict robot actions, conditioned on the imagined demonstration. We discuss details of the retrieval, training process, and video embeddings below.

\subsection{Training the Vision Encoder}
\label{sec:representation}
We align the human and robot video embeddings in three ways: visually, temporally, and at the task level, all without requiring trajectory-level correspondences. We employ unsupervised losses $\mathcal{L}_{\rm vis} (\phi)$ and $\mathcal{L}_{\rm temp} (\phi)$ for visual and temporal alignment, following prior works \cite{caron2020unsupervised, Sermanet2017TimeContrastiveNS}, and introduce an optional task alignment loss $\mathcal{L}_{\rm task} (\phi)$.

\textbf{Visual Alignment.}
To align human and robot embeddings ($\mathbf{z_R}$, $\mathbf{z_H}$), we use SwAV~\cite{caron2020unsupervised}, a self-supervised method that clusters images based on shared visual features. SwAV learns a set of $K$ prototype vectors, to which each image is assigned. The SwAV loss $\mathcal{L}_{\rm vis}(\phi)$ updates both the encoder and prototypes, aligning human and robot videos by clustering similar visual features.

\textbf{Temporal Alignment.} To align temporally adjacent frames in human and robot videos, we use Time Contrastive Loss~\cite{Sermanet2017TimeContrastiveNS}. This loss encourages embeddings of frames close in time to be similar. For each frame $z^t$, we define a positive set $\mathbf{z^{+}}$ of frames within a temporal window $w$, and a negative set $\mathbf{z^{-}}$ for frames outside this window. Using the contrastive loss $\mathcal{L}_{\rm temp}(\phi)$, we pull embeddings from the positive set closer and push negative set embeddings further apart, capturing temporal continuity across videos.

\textbf{Task Alignment.} Task-level alignment $\mathcal{L}_{\rm task} (\phi)$ is used when a small set of paired human and robot snippets is available. Unlike frame-level methods, this aligns video embeddings of the robot $\mathbf{z_R}$ and demonstrator $\mathbf{z_H}$. We compute the optimal transport distance $d(\mathbf{z_R}, \mathbf{z_H})$ to measure the similarity between two sequences of video embeddings. We then apply a contrastive learning objective (INFO-NCE~\cite{Oord2018RepresentationLW}) to pull matched embeddings closer and push different-task embeddings apart. The final task alignment loss is:

$
    \mathcal{L}_{\rm task} (\phi) = -\sum\limits_{i}\frac{\text{exp}(-d(\mathbf{z_R}^{i}, \mathbf{z_H}^{i}))}{\text{exp}(-d(\mathbf{z_R}^{i}, \mathbf{z_H}^{i})) + \sum_{j\neq i} \text{exp}(-d(\mathbf{z_R}^{i}, \mathbf{z_H}^{j}))}
$

\noindent Our final loss function for training the visual encoder $\phi$ is:
\begin{equation}
\mathcal{L} (\phi) = \lambda_{\rm vis}\mathcal{L}_{\rm vis} (\phi)+\lambda_{\rm temp}\mathcal{L}_{\rm temp} (\phi) + \lambda_{\rm task}\mathcal{L}_{\rm task} (\phi)
\vspace{-2mm}
\end{equation}

\noindent where $\lambda_{\rm task} = 0$ by default and non-zero only with access to short-horizon paired data. 


\subsection{Training the Robot Policy \label{sec:opt}}
\textbf{Training Overview.} 
Algorithm~\ref{alg:rhyme} details our approach to train robot policy $\pi_\theta$ using both robot trajectories and imagined human demonstration videos. The training process has two stages.


\emph{Stage 1: Create a Paired Dataset.} For each robot trajectory $\mathbf{\xi_R}$ and video $\mathbf{v_R}$ in $D_{\rm robot}$, we encode the robot video into embeddings $\mathbf{z_R} = \phi(\mathbf{v_R})$ using the learned video encoder $\phi$. We then retrieve imagined human embeddings $\mathbf{\hat{z}_H}$ by aligning $\mathbf{z_R}$ with demonstration snippets from the play dataset $D_{\rm play}$, through the function \texttt{Imagine-Demo}. This produces a paired dataset $D_{\rm paired}$ containing $(\mathbf{\hat{z}_H}, \mathbf{z_R}, \mathbf{\xi_R})$.

\emph{Stage 2: Train Policy on Paired Dataset.}
The policy $\pi_\theta$ is trained on the paired dataset $D_{\rm paired}$ in a hybrid fashion. For each element in $D_{\rm paired}$, we update the policy in two modes \textemdash  \emph{Mode 1:} The policy is conditioned on the robot video embeddings $\mathbf{z_R}$ to predict actions $\pi_\theta(a_t | s_t, \mathbf{z_R})$, \emph{Mode 2:} The policy is conditioned on the imagined human demonstration embeddings $\mathbf{\hat{z}_H}$ to predict actions $\pi_\theta(a_t | s_t, \mathbf{\hat{z}_H})$. 
By alternating between these two modes, the policy learns to generalize from both robot and imagined human videos, enabling it to handle execution mismatches.

\input{paper/algorithms/rhyme_algo}
\input{paper/algorithms/retrieval_algo}

\textbf{Imagining Human Demonstration Videos.} Algorithm~\ref{alg:retrieval} details the retrieval process for imagining a sequence of human embeddings. We break the robot's video into short-horizon windows and compare the embeddings with those from the play dataset, retrieving snippets with the lowest sequence-level distance. These retrieved snippets are concatenated to form an imagined long-horizon human demonstration video. The key challenge is defining a distance function $d(\mathbf{z_R}, \mathbf{z_H})$ that can handle video sequences of varying lengths. We propose two methods to compute this distance: Optimal Transport Distance and Temporal Cyclic Consistency (TCC) Distance. 



\textbf{Method 1: Optimal Transport Distance.} 
We calculate the Wasserstein distance (Optimal Transport) \cite{Peyr2018ComputationalOT} between the human and robot video embeddings, i.e., the cost of the optimal transport plan that transfers one sequence of video embeddings into another. The robot's embedding distribution is defined as $\rho_{R} = \{1/T, 1/T, \dots, 1/T\}$, and the human's embedding distribution is defined as $\rho_{H} = \{1/T', 1/T', \dots, 1/T'\}$, where $T$ and $T'$ are the lengths of the video sequences respectively. The cost function for the transport is $C \in \mathbb{R}^{T \times T'}$ where $C^{ij}$ is the cosine distance between the robot embedding $z_{R}^{i}$ and the human embedding $z_{H}^{j}$. Our goal is to find the optimal assignment $M^{*} \in \mathbb{R}^{T \times T'}$ that transports the distribution from $\rho_{R}$ to $\rho_{H}$ while minimizing the cost of the plan. Formally, we need to find $M^{*} = \argmin\limits_{M} \sum_{i}\sum_{j} C^{ij}M^{ij}$. After solving the optimal transport assignment, the distance function is the cost of the plan, i.e.,  $d(\mathbf{z_R}, \mathbf{z_H}) = \sum_{i}\sum_{j} C^{ij}M^{*ij}$. In practice, we optimize an entropy-regularized version of this problem to find an approximate solution efficiently using the Sinkhorn-Knopp algorithm~\cite{Peyr2018ComputationalOT}.

\textbf{Method 2: Temporal-Cyclic Consistency (TCC) Distance.} 
We calculate the TCC loss between human and robot videos following ~\cite{Dwibedi2019TemporalCL} which computes cycle consistency between robot video embeddings $\mathbf{z_R} = \{z_R^{1}, z_R^{2}, \dots, z_R^{T}\}$ and human video embeddings  $\mathbf{z_H} = \{z_H^{1}, z_H^{2}, \dots, z_H^{T'}\}$. For each robot frame $z_R^t$, we first compute a similarity distribution $\mathbf{\alpha}$ of $z^R_t$ with respect to the human's embeddings, to find a soft-nearest neighbor  $\Tilde{z}_H = \sum_{t'=1}^{T'} \alpha_t z^{t'}_H$. Then, $\Tilde{z}_H$ cycles back to the robot video by again computing its similarity distribution $\mathbf{\beta}$ with respect to robot video embeddings to get its soft-nearest neighbor $\Tilde{z}_R^{t} =\sum_{t=1}^{T} \beta^t z^{t}_R$. The TCC distance for a robot frame $z_R^t$ is the mean square error with its cycled-back frame $\Tilde{z}_R^{t}$ as $\ell_{\rm tcc} = ||z_R^t-\Tilde{z}_R^t||_2$. We define the video-level TCC distance function by summing over the frame-level losses $d(\mathbf{z_R}, \mathbf{z_H}) = \sum_{t=1}^T \ell_{\rm tcc}(z_R^{t})$.


We hypothesize that video retrieval using TCC distance can be inaccurate in two cases: (1) When human and robot embeddings differ due to variations in execution speed or style, leading to poor frame alignment. (2) When multiple robot embeddings correspond to a single human frame, for eg. unimanual robot tasks versus bimanual human actions.

%% file: paper/algorithms/rhyme_algo.tex
\begin{algorithm}[H]
\caption{\texttt{RHyME}: Retrieval for Hybrid imitation under Mismatched Execution}
\label{alg:rhyme}
\begin{algorithmic}
    \State {\bfseries Input:} Robot Dataset ${D}_{\rm robot}$, Human Play Dataset $D_{\rm play}$, Video Encoder $\phi(\mathbf{z}|\mathbf{v})$
    \State {\bfseries Output:} Trained Robot Policy $\pi_\theta (a|s, \mathbf{z})$
    \State Initialize Robot Policy $\pi_\theta$ 
    \State \While {\text{\upshape not converged}}
    {
        \State Get robot video and actions $\mathbf{\xi_R}, \mathbf{v_R} \sim D_{\rm robot}$
        \State Generate robot embeddings $\mathbf{z_R} = \phi(v_R)$
        \State \algcommentlight{Retrieve human embeddings}
        \State $\mathbf{\hat{z}_H} \gets \texttt{Imagine-Demo}(\mathbf{z_R}, D_{\rm play})$
        \State \algcommentlight{Hybrid Training}
        \State \For{$(s_t, a_t)$ \upshape {in} $\xi_R$}
        {
            \State \algcommentlight{Condition on imagined demo}
            \State $\texttt{Update-Policy}(a_t, \pi_\theta(s_t, \mathbf{\hat{z}_H}))$
            \State \algcommentlight{Condition on robot video}
            \State $\texttt{Update-Policy}(a_t, \pi_\theta(s_t, \mathbf{z_R}))$
        }
    }
    \State {\bfseries Return} Trained Robot Policy $\pi$
\end{algorithmic}
\end{algorithm}

%% file: paper/algorithms/retrieval_algo.tex
\begin{algorithm}[H]
\caption{\texttt{Imagine-Demo}: Retrieving Matched Human Embeddings}
\label{alg:retrieval}
\begin{algorithmic}
    \State {\bfseries Input:} Robot Embeddings $\mathbf{z_R}$, Human Play Dataset $D_{\rm play}$, Video Encoder $\phi(\mathbf{z}|\mathbf{v})$, Segment Length $K$, Distance Function $d$
    \State {\bfseries Output:} Imagined Demo $\mathbf{\hat{z}_H}$
    \State Initialize empty demo $\mathbf{\hat{z}_H} \gets \{ \}$
    \State \algcommentlight{Divide long-horizon robot sequence into short-horizon clips}
    \vspace{1mm}
    \State $Z_R = \{{z_R^{1:K}}, {z_R^{K+1:2K}}, \dots, {z_R^{T-K+1:T}}\}$
    \vspace{1mm}
    \State \For{\text{\upshape{robot segment} }${z_R^{i:i+K}} \text{ \upshape in } Z_R$}
    {
        \vspace{1mm}
        \State \algcommentlight{Find closest short-horizon clip embedding in play dataset}
        \State $\mathbf{\mathbf{\hat{z}_{H}}} \gets \argmin_{\mathbf{z_{H}} \in D_{\rm play}}{d(\mathbf{z_{H}}, {z_R^{i:i+K}}})$
        \State \algcommentlight{Extend imagined embedding sequence with retrieved demo}
        \State $\mathbf{\hat{z}_H}\text{.extend}(\mathbf{\hat{z}_{play}})$
    }
    \State {\bfseries Return} Imagined Demo $\mathbf{\hat{z}_H}$
\end{algorithmic}

\end{algorithm}


%% file: paper/experiments.tex
\section{Experiments \label{sec:experiments}}
\input{paper/figures/barplot_success}
\input{paper/figures/table_success}

\textbf{Setup.} \textbf{\texttt{Simulation:}} We evaluate our approach using the Franka Kitchen simulator~\cite{Gupta2019RelayPL}, where a 7-DOF Franka arm performs $7$ different tasks. We generate $3$ cross-embodiment video datasets each progressively increasing the embodiment and execution mismatch, which contain $580$ long-horizon robot trajectories completing a sequence of 4 tasks and a bank of cross-embodiment demonstrator play data ($>$\;3 hours) for training our models. First, in \textbf{\textsc{Sphere-Easy}}, we replace the robot's visual rendering with a sphere following the gripper’s position, creating a visual gap between robot and demonstrator. Second, in \textbf{\textsc{Sphere-Medium}}, we introduce manipulation style mismatches such as the robot dragging an object while the demonstrator lifts and carries it. Finally, in \textbf{\textsc{Sphere-Hard}}, we create a further divergence where the demonstrator performs two tasks simultaneously, similar to how humans use two hands. Fig.~\ref{fig:dataset_barplot} (left) illustrates the cross-embodiment datasets. 
\textbf{\texttt{Realworld:}} We use a 7-DOF Franka arm to perform $4$ different tasks. We train our models on $40$ long-horizon robot trajectories completing a sequence of 3 tasks and $\sim$15 minutes of human play data, holding out unseen compositions of 3 test tasks.


\textbf{Baselines.} \texttt{XSkill}~\cite{xu2023xskill} simply conditions on robot videos during train-time, and uses its shared representation space to zero-shot generalize to inputs of human videos at test-time. \texttt{OraclePairing}~\cite{Jang2022BCZZT,Jain2024Vid2RobotEV} is the gold-standard approach, assuming an oracle pairs human demonstrations with robot trajectories, enabling conditioning on the human at train time. Our approach, \texttt{RHyME}, finds a middle ground. Without pairing, it \textit{imagines} human videos that perform the same tasks as a robot trajectory (Section \ref{sec:opt}) by exploiting sequence-level correspondences. We compare two variants of our algorithm, \texttt{RHyME-TCC} and \texttt{RHyME-OT}, which differ in their distance functions used for retrieval. In the realworld, we use \texttt{XSkill}~\cite{xu2023xskill} as a baseline for \texttt{RHyME-OT}. We also show how vision representations can be improved using short-horizon robot-demonstrator task pairs. 

\textbf{Evaluation and Success Metrics.} 
At test-time, we provide the robot policy with long-horizon human videos as prompts. \textbf{\textit{Simulation:}} We evaluate one-shot imitation performance across 20 different demonstrator videos, rolling out the robot policy from the last 5 model checkpoints, yielding 100 total trials per dataset. We measure \textit{Task Recall}, which assesses recall by counting the successfully completed tasks shown in the demonstrator's video, and \textit{Task Imprecision}, which measures imprecision and reports the percentage of tasks the robot attempts incorrectly—those not specified in the demonstrator's video. \textbf{\textit{Realworld:}} We evaluate performance across 30 different human videos (20 seen, 10 unseen). We break down \textit{Task Recall} into two metrics: \textit{Task Attempts} and \textit{Task Completions}, which measure (a) the robot's ability to attempt tasks specified by the human video and (b) the low-level control policy's ability to fully complete the tasks.

\input{paper/figures/tsne}

\textbf{\textit{Q1.} How does performance vary across different levels of execution mismatch?} As the cross-embodiment demonstrator's execution deviates further from those of the robot, policies trained with our framework \texttt{RHyME} consistently outperform \texttt{XSkill} (Fig.~\ref{fig:dataset_barplot}), with the largest gap in the bimanual demonstrator setting \textbf{\textsc{Sphere-Hard}} ($53\%$ vs $1\%$). The \texttt{OraclePairing} baseline serves as an upper bound on performance.

Fig.~\ref{fig:tsne} (left) investigates this trend by probing the visual representations of the video encoder $\phi$, common across policies. We plot the image embeddings of the robot and demonstrator across three different tasks using a t-SNE plot. As execution mismatch increases, the robot and demonstrator embeddings become less clustered by task, supporting \texttt{XSkill}'s inability to zero-shot transfer to demonstrator embeddings at test-time in \textbf{\textsc{Sphere-Hard}}. \texttt{RHyME} algorithms overcome this problem and successfully retrieve the correct demonstration videos at train-time by reasoning over sequences of embeddings. 
However, \texttt{RHyME-OT} outperforms \texttt{RHyME-TCC} across all three datasets in both metrics, suggesting inaccurate train-time retrievals with TCC.

\textbf{\textit{Q2.} How does \texttt{RHyME} perform on real kitchen tasks when prompted with human videos?} With natural visual and execution mismatches between human and robot videos, \texttt{RHyME} consistently outperforms \texttt{XSkill} when prompted with both seen and unseen human prompt videos (Fig. \ref{fig:realworld-tsne} Right). We observe marginal benefits in \textit{Task Attempts} and \textit{Task Completions} when faced with the compositions of the tasks seen, but record significant improvements in both metrics (83\% vs. 50\% and 67\% vs. 33\%, respectively) in the unseen setting to which our framework aims to generalize. 


\input{paper/figures/qualitative_retrievals}

\textbf{\textit{Q3.} How does video retrieval using Optimal Transport and TCC impact policies at test-time? } As task embedding clusters deviate due to execution mismatches, we observe inaccuracies in TCC retrievals: in \textbf{\textsc{Sphere-Hard}} (Fig.~\ref{fig:tsne} (right)) when both clips complete the same two tasks, a bimanual task embedding lies in between two robot task clusters which results in cycling-back to the incorrect robot frame, leading to high task imprecision. \texttt{RHyME-OT} performs strictly better across datasets (Fig.~\ref{fig:dataset_barplot}). The key reason for this performance difference is that optimal transport computes distances by matching videos across a sequence of embeddings. Fig.~\ref{fig:ot_plot} visualizes the cost of the optimal transport plan between prompt robot clips and demonstration videos in the hard \textbf{\textsc{Sphere-Hard}} dataset. Comparing a robot clip doing two tasks (e.g. kettle and light), the transport cost across assignments is minimum only when compared to the demonstrator performing those same two tasks. TCC, on the other hand, attempts to establish one-to-one correspondences between the robot and demonstration frames, which are lacking in this dataset.

\textbf{\textit{Q4.} Where does \texttt{RHyME} succeed, and what are common failure modes of other methods?} We visualize the vision embeddings using t-SNE (Fig.~\ref{fig:realworld-tsne} Left). We find that image embeddings in the realworld are generally clustered by task, but tend to deviate during the \textit{Light Switch} task. Consequently, we observed that \texttt{XSkill} never attempts the task when prompted with human embeddings. On the other hand, in unseen settings, \texttt{RHyME} always attempts the \textit{Light Switch} task and completed it 9 out of 10 times. The Optimal Transport retrieval (Sec.~\ref{sec:approach}) used to imagine the paired dataset recognizes can correctly match human and robot clips completing the same task by reasoning over the distribution of embeddings rather than relying on perfect embedding alignment, so \texttt{RHyME} is able to accurately pair action labels with imagined human videos at train time and obtain better performance (Fig.~\ref{fig:realworld-tsne} Right).

\input{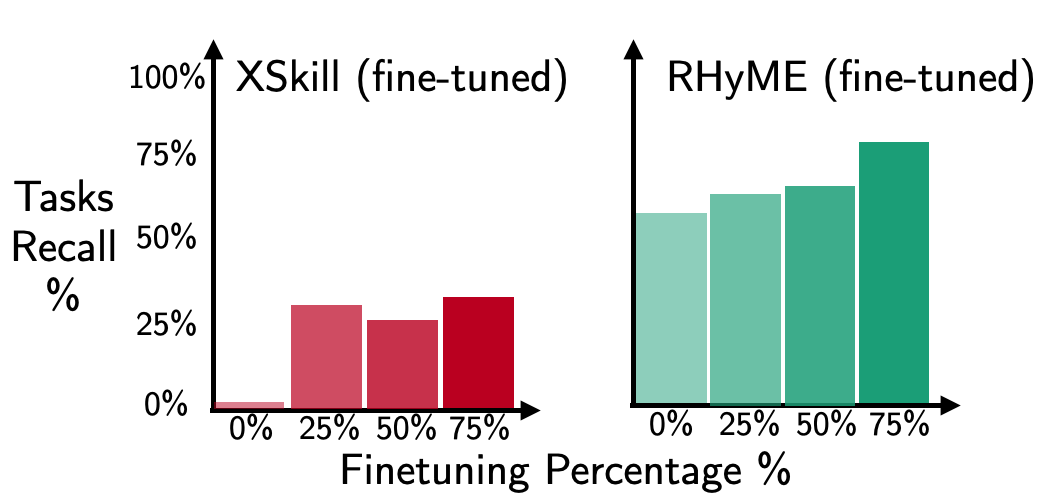}

\textbf{\textit{Q5.} Does fine-tuning visual representations with task-equivalent pairs improve one-shot imitation?} Following Section~\ref{sec:representation}, we assume access to short-horizon task pairings across embodiments and apply $L_{task} (\phi)$ on vision representations in the \textbf{\textsc{Sphere-Hard}} setting. We find that encouraging induced distributions over embeddings to be similar lifts the performance of both \texttt{XSkill} and \texttt{RHyME-OT}  (Fig.~\ref{fig:finetune}), and scales up with more paired clips. Ultimately, comparing induced distributions over embeddings with optimal transport is a beneficial design choice for matching clips at a task-level in the face of execution mismatches, as \texttt{RHyME-OT (0\% fine-tuned)} still significantly outperforms \texttt{XSkill (fine-tuned)}.

\input{paper/figures/generalization}

%% file: paper/figures/barplot_success.tex
\begin{figure*}[t!]
    \centering
    \includegraphics[width=0.9\linewidth]{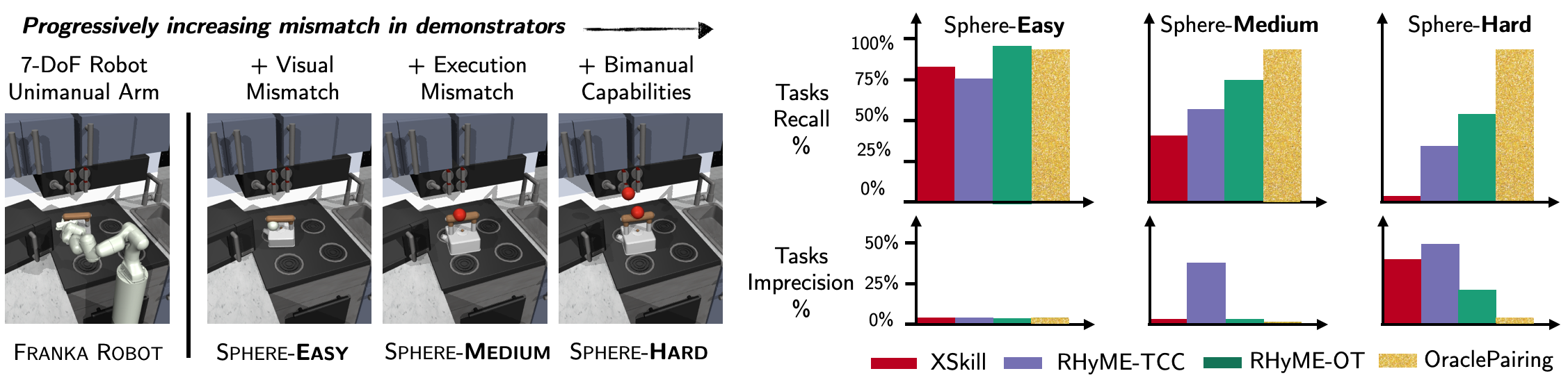}
    \vspace{-2mm}
    \captionsetup{width=\textwidth}
    \caption{\textbf{Performance on Mismatched Execution Datasets.} 
    We present results on three datasets (left). As the demonstrator's execution deviates further from those of the robot, policies trained with our framework \texttt{RHyME} consistently outperforms \texttt{XSkill} measured by task recall and imprecision rates.}  
    \label{fig:dataset_barplot}
\end{figure*}

%% file: paper/figures/table_success.tex
\begin{figure*}[t!]
    \centering
    \includegraphics[width=0.9\linewidth]{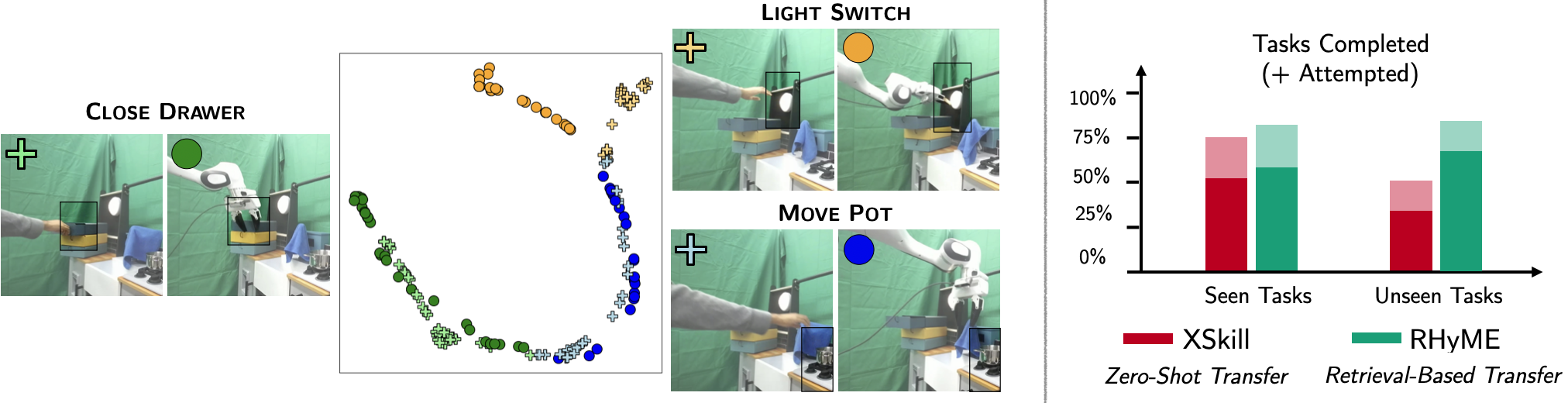}
    \captionsetup{width=\textwidth}
    \vspace{-2mm}
    \caption{\textbf{Realworld Results}. (Left) \textbf{Task Embeddings:} We use t-SNE to visualize cross-embodiment latent embeddings from the human and robot completing three tasks. (Right) \textbf{Task Completion:} We compare the performance of \texttt{RHyME} with \texttt{XSkill} on seen and unseen long-horizon tasks specified by human prompt videos. Opaque segments indicate \textit{Task Completion} rate, and augmented transparent bars indicate \textit{Task Attempt} rate.}  
    \label{fig:realworld-tsne}
\label{fig:realworld-tsne}
\end{figure*}

%% file: paper/figures/tsne.tex

\begin{figure*}[t!]
    \centering
    \includegraphics[width=0.9\linewidth]{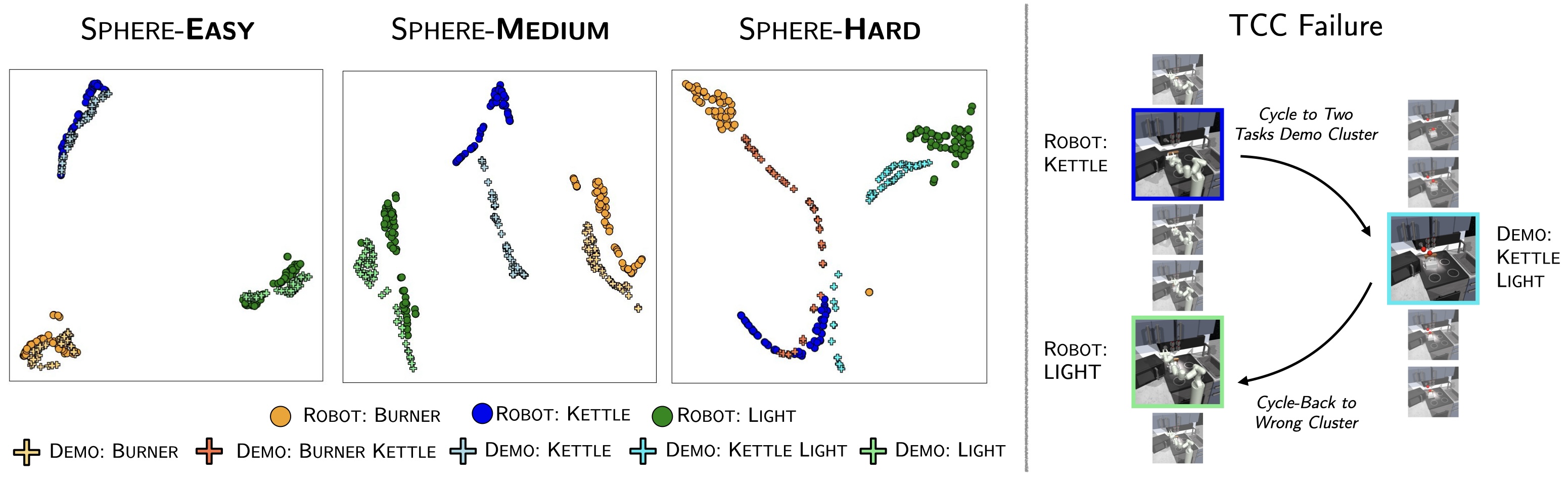}
    \vspace{-1mm}
    \captionsetup{width=\textwidth}
    \caption{\textbf{Cross-Embodiment Vision Embeddings.} (Left) \textbf{Visualizing task embeddings.} We use t-SNE to visualize cross-embodiment latent embeddings generated by robot and demonstrator when executing different tasks on all three datasets. 
    (Right) \textbf{TCC Failure Example:} The robot and video clip are equivalent, but specific frames have high TCC losses. For example, a frame showing the robot performing the `kettle' action has a high loss due to its nearest neighbor in the video performing both `kettle' and `light' actions. This frame cycles back to the robot performing `light', which is mismatched.}  
    \vspace{-5mm}
    \label{fig:tsne}
\end{figure*}


%% file: paper/figures/qualitative_retrievals.tex
\begin{figure}[t!]
    \centering
    \includegraphics[width=0.8\linewidth]{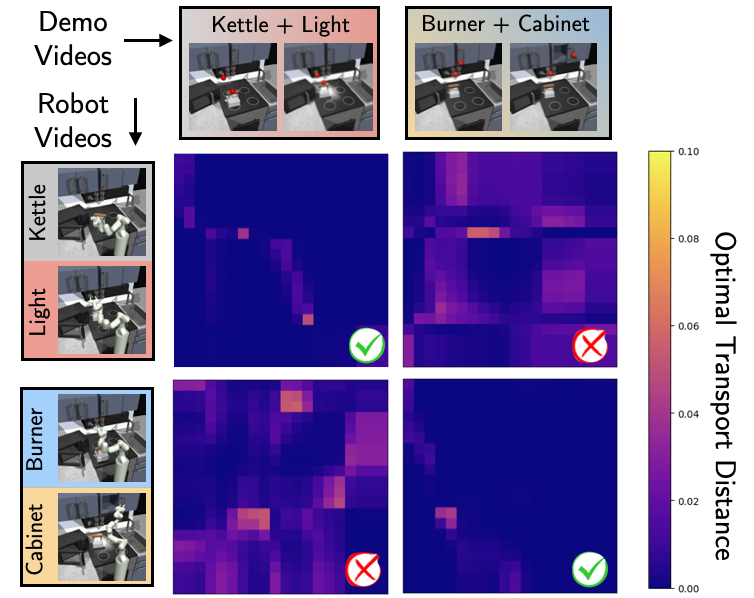}
    \vspace{-2mm}
    \captionsetup{width=\linewidth}
    \caption{\textbf{\textit{Optimal Transport Distances.}}
    We measure the similarity between robot and demonstrator videos on the \textbf{\textsc{Sphere-Hard}} dataset by computing the cost of the Optimal Transport (OT) plans. The sum over the entire transport cost matrix costs yields the distance between videos. OT costs are lowest when tasks are the same between videos (highlighted by a tick).
    }
    \label{fig:ot_plot}
\end{figure}


%% file: paper/figures/pairing_performance.tex

\begin{figure}[t!]
    \centering
    \includegraphics[width=0.8\linewidth]{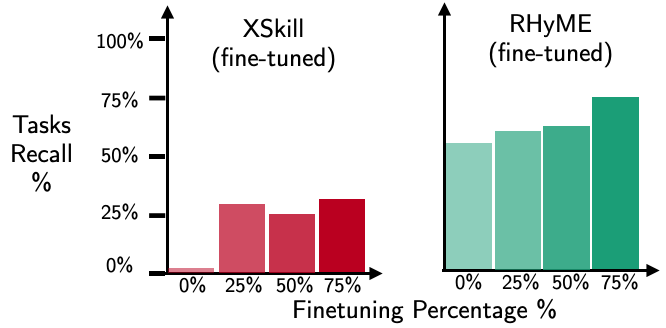}
    \vspace{-2mm}
    \captionsetup{width=\linewidth}
    \caption{\textbf{Performance Improves by Pairing Skills on \textsc{Sphere-Hard}.} For both non-retrieval and retrieval-based methods, performance improves when fine-tuned when the visual encoder is finetuned short-horizon robot-demonstrator snippet pairs using a contrastive optimal transport loss.
    }  
    \vspace{-8mm}
    \label{fig:finetune}
\end{figure}


%% file: paper/discussion.tex
\section{Discussion and Limitations}


This work addresses the challenge of one-shot imitation in the presence of \textit{mismatched execution} by the demonstrator. We propose \texttt{RHyME}, a novel framework that leverages task-level correspondences to bridge frame-level visual disparities between the robot and the demonstrator, enabling the learning of a video-conditioned policy without paired data.

\textbf{Limitations.} While the exact test-time task compositions are unseen during training, our method relies on transitions between task pairs in the robot dataset to learn transition actions. This limits the ability to learn entirely new task sequences. We note that our method still generalizes well to new compositions when such transitions are present. 


